\documentclass{article}
\usepackage{ijcai25}

\usepackage{times}
\usepackage{soul}
\usepackage{url}
\usepackage[hidelinks]{hyperref}
\usepackage[utf8]{inputenc}
\usepackage[small]{caption}
\usepackage{graphicx}
\usepackage{amsmath}
\usepackage{amsthm}
\usepackage{booktabs}
\usepackage{algorithm}
\usepackage{algpseudocode}
\usepackage[switch]{lineno}
\usepackage{multirow}
\usepackage{array}
\usepackage{amsfonts}
\usepackage{cleveref}

\usepackage[T1]{fontenc}    
\usepackage{nicefrac}       
\usepackage{microtype}      
\usepackage{xcolor}         

\usepackage{epsfig}
\usepackage{amssymb}
\usepackage{enumitem}
\usepackage[font=small]{caption}
\usepackage{subcaption}
\usepackage{array}
\usepackage{cleveref}
\Crefformat{figure}{Fig.~#2#1#3}                           %
\Crefname{subfigure}{Fig.}{Figs.}
\Crefname{figure}{Fig.}{Figs.}
\Crefformat{table}{TABLE~#2#1#3}                           %

\renewcommand{\vec}[1]{\boldsymbol{#1}}
\renewcommand{\L}{\vec{L}}
\newcommand{\R}{\vec{R}}
\newcommand{\I}{\vec{I}}
\newcommand{\minisection}[1]
{\noindent{\textbf{#1}}.}

\title{Low-Light Video Enhancement via Spatial-Temporal Consistent Decomposition}

\author{
Xiaogang Xu$^{1,2}$
\and Kun Zhou$^3$
\and Tao Hu$^4$
\and Jiafei Wu$^{5}$$^*$
\and Ruixing Wang$^{6}$\footnote{Corresponding authors.}
\and Hao Peng$^7$
\and Bei Yu$^1$
\affiliations
$^1$The Chinese University of Hong Kong\\
$^2$Zhejiang University \\
$^3$The Chinese University of Hong Kong (Shenzhen) \\
$^4$PICO, Bytedance \\
$^5$The University of Hong Kong\\
$^6$DJI Technology Co., Ltd. \\
$^7$Zhejiang Normal University\\
\emails
xiaogangxu00@gmail.com, kunzhou@link.cuhk.edu.cn, yihouxiang@gmail.com, \\
jcjiafeiwu@gmail.com, ruixing0406@gmail.com, hpeng@zjnu.edu.cn, byu@cse.cuhk.edu.hk \\
}

\begin{document}

\maketitle

\begin{abstract}
Low-Light Video Enhancement (LLVE) seeks to restore dynamic or static scenes plagued by severe invisibility and noise. 
In this paper, we present an innovative video decomposition strategy that incorporates view-independent and view-dependent components to enhance the performance of LLVE. We leverage dynamic cross-frame correspondences for the view-independent term (which primarily captures intrinsic appearance) and impose a scene-level continuity constraint on the view-dependent term (which mainly describes the shading condition) to achieve consistent and satisfactory decomposition results.
To further ensure consistent decomposition, we introduce a dual-structure enhancement network featuring a cross-frame interaction mechanism. By supervising different frames simultaneously, this network encourages them to exhibit matching decomposition features.
This mechanism can seamlessly integrate with encoder-decoder single-frame networks, incurring minimal additional parameter costs. 
Extensive experiments are conducted on widely recognized LLVE benchmarks, covering diverse scenarios. Our framework consistently outperforms existing methods, establishing a new SOTA performance.
\end{abstract}

\section{Introduction}
\label{sec:intro}

Low-light enhancement aims to enhance underexposed images and videos captured in low-light conditions~\cite{xu2022snr,wang2021sdsd}, improving their visual quality while reducing noise. 
This technique can be applied in wide-ranging applications, such as portrait photography on mobile devices~\cite{ignatov2017dslr,hasinoff2016burst}, nighttime face recognition~\cite{ma2022toward,wang2022unsupervised} and vehicle detection~\cite{fu2023dancing,wu2022edge}.
The key challenge when enhancing videos is the need for consistent enhancement results across corresponding locations in different frames, which vary both spatially and temporally.
Moreover, there is a scarcity of real-world, high-quality, spatially aligned video pairs for dynamic scenes, resulting in limited generalization capabilities for unseen videos with varying conditions.

\begin{figure}[tb!]
    \centering
    \includegraphics[width=1.0\linewidth]{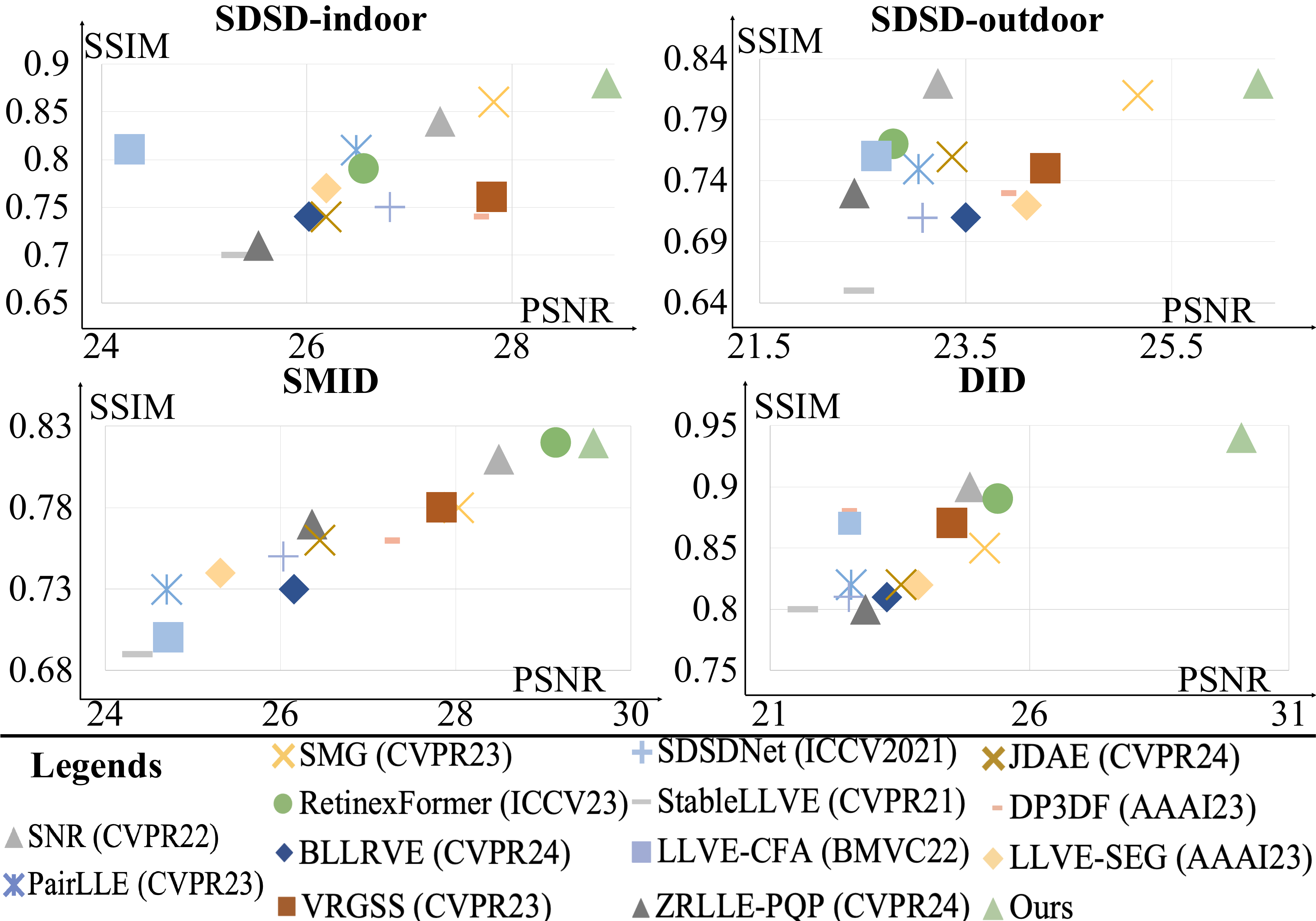}
    \caption{
        Our proposed LLVE method {\em consistently\/} achieves SOTA performance on {\em different LLVE datasets involving various scenes\/} with the same network architecture. 
    }
    \label{fig:teaser-com}
\end{figure}

One solution is to incorporate parameterized enhancement models with physical significance into the enhancement process, reducing the reliance on training data.
According to the linear Lambertian model for intrinsic image decomposition~\cite{Lambertian,narihira2015direct}, an observed image $\I$ can be expressed as the element-wise product ($\otimes$) of its albedo component and shading component.
The albedo part is normally view-independent, capturing the intrinsic appearance, whereas the shading component is commonly view-dependent, influenced by lighting conditions and the surface normal's direction.
Inspired by the Lambertian model, we propose that a frame in the LLVE task can similarly be decomposed as $\I = \L \otimes \R$, where $\L$ represents the view-dependent term, and $\R$ corresponds to the view-independent component.

In this paper, we introduce an innovative approach to attain our decomposition for normal-light video outputs, ensuring consistency in both spatial and temporal dimensions. 
We leverage cross-frame correspondences and real-world physical continuity constraints to achieve this decomposition without the need for explicit supervision. Furthermore, to enhance the consistency of the decomposition, we've developed an efficient dual-structure enhancement network featuring a novel cross-frame interaction mechanism.

Our approach aims to predict both the $\R$ and $\L$ of normal-light images from the provided low-light inputs.
It is important to note that the decomposition of view-independent and view-dependent terms is not unique, and our goal is to implement a suitable decomposition that enhances the performance of LLVE.
In this paper, we aim to design appropriate priors directly from the video data to implement the decomposition.
We recognize that $\R$ should represent the physical properties of objects, which should remain consistent across various observation perspectives. To achieve this, we first calculate corresponding locations in videos and utilize these computed spatial-temporal correspondences, along with uncertainty values, to link the predictions of $\R$ across different frames.
Conversely, the terms of $\L$ are subject to a spatial smoothness loss to model real-world physical constraints (since the lighting field is continuous in practice). This approach allows us to establish a temporal-spatial consistency constraint in the enhanced videos, without requiring additional supervision.
\textit{Note that we cannot guarantee the decomposition results are strictly view-independent or view-dependent, as this is not a 3D method. Our goal is to apply appropriate constraints to approximate desired decomposition properties, while focusing on optimizing the final LLVE performance with our approach.}

Additionally, we introduce a dual-network structure to enforce the consistency of feature representations across frames, facilitating the consistent synthesis of decomposition in turn.
Two frames are used as inputs, and their features are propagated within the designed Cross-Frame Interaction Module (CFIM).
CFIM differs from other similar architectures in the restoration task~\cite{lv2023unsupervised} in terms of the fusion manner.
CFIM facilitates interaction through a combination of long-range cross-frame attention computation and short-range channel-spatial fusion, ensuring the accurate and comprehensive merging of features from both global and local perspectives. 
Moreover, CFIM selectively incorporates knowledge from randomly chosen neighboring frames into the backbone (i.e., the part without feature propagation) during training. In this way, CFIM help the backbone in learning how to adaptively utilize knowledge from various cross-frame features, resulting in a more robust backbone.

We performed experiments on various widely recognized video enhancement datasets. Our results, both quantitative and qualitative, consistently demonstrate the effectiveness and state-of-the-art (SOTA) performance of our framework, as shown in \Cref{fig:teaser-com}.
Furthermore, we conducted a large-scale user study involving 100 participants, which showcased the superiority of our results in terms of human subjective perceptions.
In summary, our contributions are three-fold.
\begin{itemize}
	\item We introduce a novel canonical form for LLVE that predicts spatial-temporal consistent decomposition with view-independent and view-dependent terms for normal-light outputs.
    This decomposition strategy leverages spatial-temporal correspondences and continuity.
	\item We design a new LLVE network that facilitates interaction among the features of different frames and fits our consistent decomposition strategy.
	\item We conduct extensive experiments on public datasets, illustrating the effectiveness of our proposed framework.
\end{itemize}

\section{Related Work}
\label{sec:formatting}

\noindent\textbf{Low-Light Image Enhancement.}
To enhance the quality of a low-light video, image enhancement methods can be applied on a frame-by-frame basis. In recent years, learning-based Low-Light Image Enhancement (LLIE) techniques~\cite{jiang2021enlightengan,yang2021band} have made significant advancements, with a primary focus on supervised approaches.

\noindent\textbf{Low-Light Video Enhancement.}
In addition to the need for LLIE, there is a growing demand for video enhancement, considering the widespread use of videos as a popular data format on the internet and in photographic equipment. Various approaches have been proposed in this context~\cite{chen2019seeing,triantafyllidou2020low,ye2023spatio,lv2023unsupervised,xu2023deep,fu2023dancing}. 
Liu et.al.~\cite{liu2023low} and Liang et.al.~\cite{liang2023coherent} used prior event information to learn enhancement mapping for brightening videos. Xu et.al.~\cite{xu2023deep} designed a parametric 3D filter tailored for enhancing and sharpening low-light videos. Recently, Fu et.al.~\cite{fu2023dancing} introduced a video enhancement method called LAN, which iteratively refines illumination and adaptively adjusts it. However, it's important to note that LAN lacks an explicit constraint for maintaining consistent reflectance and illumination decomposition.

Several datasets for video enhancement, encompassing both static~\cite{chen2019seeing,jiang2019learning,wang2019enhancing,triantafyllidou2020low} and dynamic motions~\cite{wang2021sdsd,fu2023dancing}, have been introduced. In this paper, we introduce a novel LLVE method, designed to enhance effects on these datasets. Our decomposition method explicitly and consistently models view-dependent and -independent for all frames.

\begin{figure*}[tb]
    \centering
    \includegraphics[width=.8\linewidth]{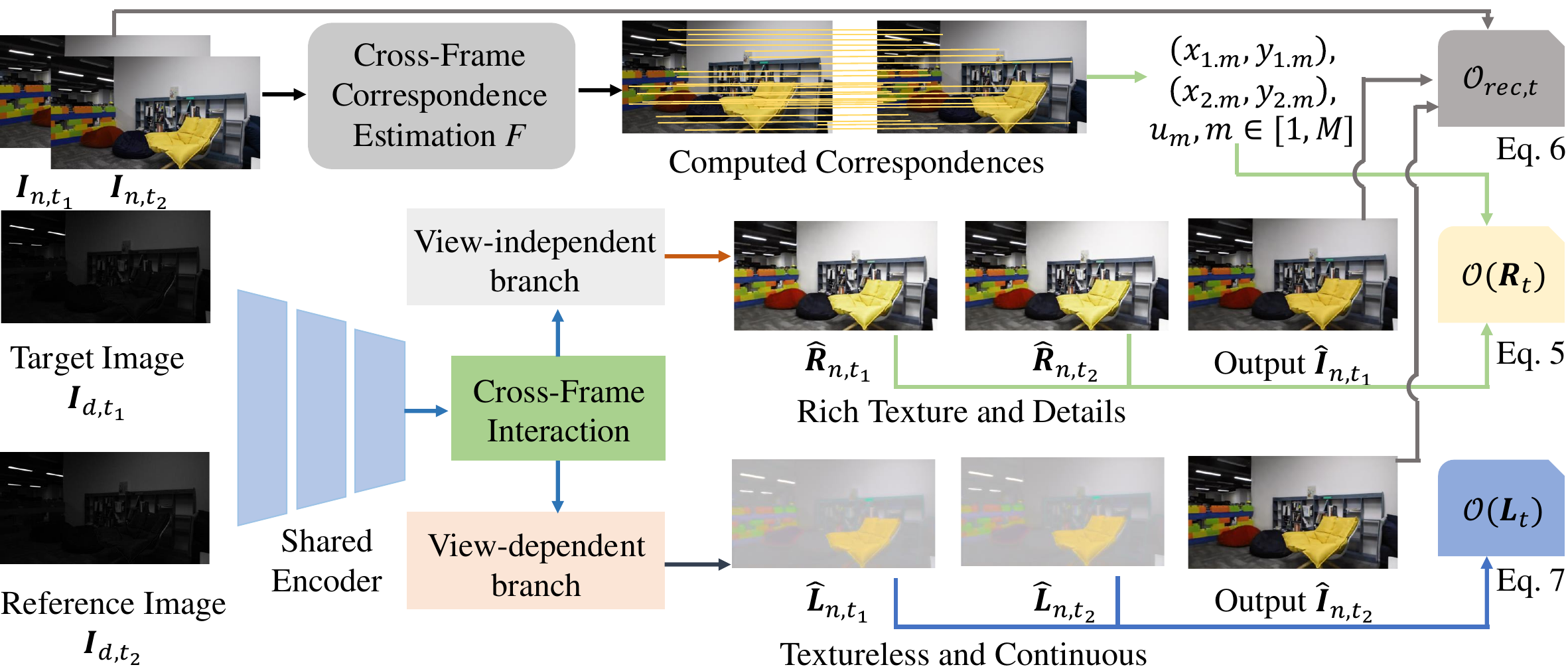}
    \caption{
        Our framework offers a comprehensive solution that explicitly and consistently models the view-independent and view-dependent decomposition of enhanced normal-light outputs across different frames.
        To achieve this, we enforce consistent features in the view-independent terms across different frames by leveraging computed correspondences in the temporal dimension of videos ($\mathcal{O}(\R_t)$). Simultaneously, we ensure that the view-dependent terms exhibit a spatially continuous distribution ($\mathcal{O}(\L_t)$), aligning with real-world scenarios. Furthermore, our network incorporates cross-frame interaction and simultaneous supervision of different frames within a video, encouraging consistent features for these frames derived from one video. For a more detailed visual representation, please refer to \Cref{fig:framework2}. 
    }
    \label{fig:framework}
\end{figure*}

\section{Method}

\subsection{Decomposition Model}
\label{sec:1}

\minisection{Motivation} 
According to the linear Lambertian model~\cite{narihira2015direct}, an observed image $\I$ can be formulated as the element-wise product of its albedo and shading component. Here, the albedo part is view-independent, capturing the intrinsic appearance, while the shading component is view-dependent.
Inspired by this, when presented with an image $\I$ (in this paper, we denote the low-light image as $\I_d$ and the normal-light image as $\I_n$, with $d$ and $n$ serving as subscript abbreviations),
we assume its decomposition into view-dependent part $\L$ and view-independent part $\R$, as
\begin{equation}
    \I=\L \otimes \R,
\end{equation}
where $\otimes$ denotes the element-wise multiplication, $\L$ and $\R$ can be formulated for different channels, i.e., the channel number is 3 for the sRGB domain.
In this context, $\L$ mainly describes the light intensity of objects, which is expected to exhibit piece-wise continuity.
On the other hand, $\R$ primarily represents the physical properties of the objects, encompassing textures and details observed in the data $\I$.
If we obtain $\L$ and $\R$ in the normal-light conditions for a given input low-light image, the target normal-light image can be acquired accordingly.
The common objective can be summarized as
\begin{equation}
	\mathcal{O}=\Vert \I-\L\otimes \R \Vert + \mathcal{O}(\L)+\mathcal{O}(\R),
	\label{eq:image}
\end{equation}
where $\mathcal{O}(\L)$ and $\mathcal{O}(\R)$ denote the constraints for $\L$ and $\R$ (we will introduce our designed new constraints in Sec.~\ref{sec:2}).
When applying this model to sequential video data, the objective related to the decomposition of video can be written as
\begin{equation}
	\begin{aligned}
	&\mathcal{O}_v=\mathbb{E}_{t=1:T} [\mathcal{O}_{rec,t} + \mathcal{O}(\L_t)+\mathcal{O}(\R_t)],\\
	&\mathcal{O}_{rec,t}=\Vert \I_t-\L_t\otimes \R_t \Vert,
	\end{aligned}
	\label{eq:video}
\end{equation}
where $\mathbb{E}$ is the average operation, $T$ is the number of frame, and $\L_t$/$\R_t$ is the decomposition output at the time index of $t$. Compared with \Cref{eq:image}, we need to simultaneously guarantee the decomposition quality and the temporal consistency.

\minisection{Video Data Guide the Decomposition by Themselves}
We have discovered that video data itself can play a crucial role in facilitating the achievement of our decomposition predictions for normal-light outputs, thus obviating the need for external priors typically used in low-light enhancement. By establishing correspondences among different frames, we can impose specific constraints: the $\R_t$ of each frame should faithfully represent the intrinsic texture of objects within the target scene, remaining consistent regardless of changes in viewpoint. Likewise, the $\L_t$ of each frame should exhibit the desired continuity consistent with real-world physical properties. For further elaboration, please refer to \Cref{sec:2}.

\minisection{Problem Formulation}
We adopt the supervised setting. Given a clip of low-light data $\I_{d, t}, t\in[1, T]$, there is a paired normal-light data $\I_{n, t}, t\in[1, T]$. 
\textit{We aim to directly obtain the decomposition of $\I_{n, t}$ from $\I_{d, t}$ using network $f$}, as
\begin{equation}
	\hat{\L}_{n, t}, \hat{\R}_{n, t}= f(\I_{d, t}), \hat{\I}_{n, t}=\hat{\L}_{n, t}\otimes \hat{\R}_{n, t},
\end{equation}
where $\hat{\L}_{n, t}$ and $\hat{\R}_{n, t}$ are the estimated targets, and $\hat{\I}_{n, t}$ is the predicted enhancement result.
Our framework is shown in \Cref{fig:framework} that will be introduced in the following sections.

\subsection{Spatial-Temporal Consistent Decomposition}
\label{sec:2}

\minisection{Motivation} 
Video can be conceptualized as static/dynamic 3D multi-view data~\cite{wang2023tracking,wang2023lighting}. Prior research has demonstrated that the multi-view data itself can be harnessed to decompose the view-dependent and view-independent elements, which respectively encapsulate the intrinsic texture and view-altering illuminations~\cite{wang2023lighting}. Thus, once we establish correspondence relationships among $\I_{d,t}$ for all $t$, we can apply the view-independent constraint to derive $\hat{\R}_{n, t}$, as it represents the intrinsic properties of the target scene. 
Subsequently, obtaining $\hat{\L}_{n, t}$ becomes achievable through utilizing the inherent reconstruction loss, in conjunction with adherence to the continuity assumption.

\minisection{Implementation of Correspondences}
Obtaining correspondences for $\I_{d,t}$, $t\in[1, T]$, can be a challenging task due to the presence of various degradations, including visibility issues and noise. Fortunately, we have access to corresponding normal-light data, which significantly aids in establishing these correspondences, as illustrated in \Cref{fig:framework}, where $\I_{d,t}$ and $\I_{n,t}$ are pixel-wise aligned.

Given two frames $\I_{n,t_1}$ and $\I_{n,t_2}$, correspondences can be determined using the prediction network $F$~\cite{edstedt2023dkm}. We assume the detection of $M$ correspondences, denoted as ${ c_m=(x_{1,m}, y_{1,m}, x_{2,m}, y_{2,m}), m\in[1, M]}$. Each correspondence, represented by $c_m$, consists of four coordinates: $(x_{1,m}, y_{1,m})$ represents the pixel coordinate in $\I_{n,t_1}$, while $(x_{2,m}, y_{2,m})$ signifies the coordinate in $\I_{n,t_2}$. Furthermore, each correspondence is associated with an uncertainty value $u_m$, determined through a data-driven approach.

 \begin{figure}[tb!]
     \centering
     \includegraphics[width=1.0\linewidth]{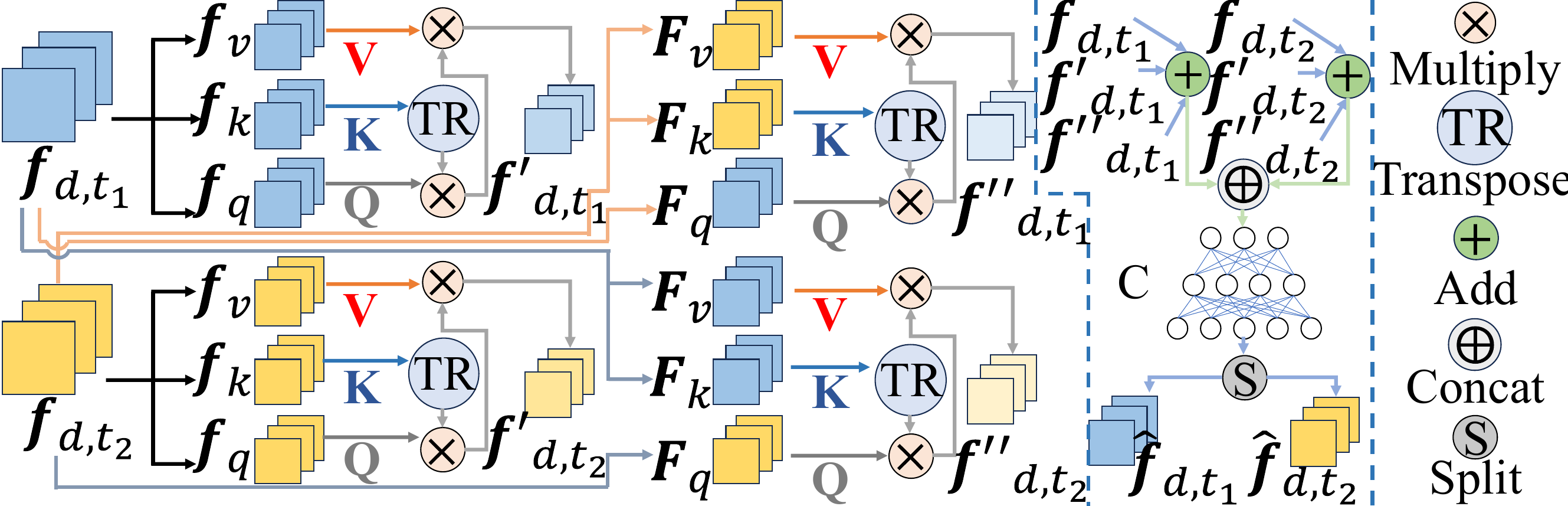}
     \caption{The lightweight cross-frame interaction mechanism (CFIM) propagates different frames' features, along with cross-frame attention and spatial-channel fusion. 
     Cross-frame interaction can be employed in the deep feature space of arbitrary single-image encoder-decoder frameworks, and we choose U-Net here.
     }
     \label{fig:framework2}
\end{figure}

\minisection{Constraints Formulation}
To satisfy the constraint that the corresponding pixels in $\R_t$ are matched, we have $\mathcal{O}(\R_t)$ as
\begin{equation}
\small
	\mathcal{O}(\R_{t_1,t_2})=\mathop{\mathbb{E}}\limits_{m\in[1, M]} u_m\Vert \hat{\R}_{n, t_1}[x_{1,m}, y_{1,m}]- \hat{\R}_{n, t_2}[x_{2,m}, y_{2,m}] \Vert .
	\label{eq:1}
\end{equation}
The reconstruction item in \Cref{eq:video} can be denoted as 
\begin{equation}
	\mathcal{O}_{rec,t}=\Vert \I_{n, t}-\hat{\L}_{n, t}\otimes \hat{\R}_{n, t} \Vert.
	\label{eq:2}
\end{equation}
Although the view-dependent part can be obtained via the reconstruction constraint, 
We further incorporate a continuity constraint to align with real-world properties. 
For a pixel in the video frame, denoted as $p$, we introduce a spatial continuity objective on the predicted $\hat{\L}_{n,t}$ as the following loss function
	\begin{equation}
		\mathcal{O}(\L_t)=\mathbb{E}_t [v_t^p \times [\partial_x \hat{\L}_{n,t}(p)]^2 + u_t^p \times [\partial_y \hat{\L}_{n,t}(p)]^2],
		\label{eq:3}
	\end{equation}
	where $\partial_x$ and $\partial_y$ are partial derivatives in horizontal and vertical directions, respectively. $v_t^p$ and $u_t^q$ are spatially-varying smoothness weights, calculated as
	\begin{equation}
 \small
		v_t^p=(\Vert \partial_x \boldsymbol{U}_{t}(p) \Vert + \Delta )^{-1}, u_t^p=(\Vert \partial_y \boldsymbol{U}_t(p) \Vert + \Delta)^{-1},
		\label{eq:smooth}
	\end{equation}
	where $\boldsymbol{U}_t$ represents the logarithmic transformation of $\I_{d,t}$, and $\Delta$ is a small constant (set to 0.0001) used to avoid division by zero. 
	The diagram is shown in \Cref{fig:framework}.

	\subsection{Dual Network for Consistent Decomposition}
	\label{sec:3}

	\minisection{Overview and Motivation} 
	To enhance the consistency of decomposition, it's crucial for various frames within a video to mutually share features. By incorporating shared features and simultaneous supervision, the features used for synthesizing decomposition parameters across different frames can exhibit greater consistency. When the video is enhanced frame by frame without feature propagation, the decomposition results tend to be suboptimal due to the inherently challenging nature of maintaining consistency. This observation is empirically validated in our ablation study. Current LLVE methods primarily leverage multiple-frame inputs to facilitate propagation~\cite{wang2021sdsd,fu2023dancing}. However, these strategies come with a notable cost in terms of aligning features.

	In this paper, we introduce an efficient strategy for propagation using a dual approach. This approach entails loading two frames, specifically target frame $\I_{d, t_1}$ and reference frame $\I_{d, t_2}$ (where $\I_{d, t_2}$ can be randomly selected from the temporal neighbors of $\I_{d, t_1}$ during training, and is set as the closest frame of $\I_{d, t_1}$ during inference),
    propagating features at the deepest layer of the network through our cross-frame interaction (as depicted in \Cref{fig:framework2}), and concurrently supervising these dual outputs to guarantee consistency.

 \minisection{Feature Propagation via Spatial-varying Fusion}
	Suppose the feature of $\I_{d, t_1}$ and $\I_{d, t_2}$ are denoted as $\boldsymbol{f}_{d, t_1}$ and $\boldsymbol{f}_{d, t_2}$ which is extracted from the same encoder $M_E$ in the network $f$.
    To complete the propagation, we initially employ a long-range cross-frame attention operation in the feature space. A traditional attention operation~\cite{vaswani2017attention} typically consists of the query vector $\boldsymbol{Q}$, the key vector $\boldsymbol{K}$, and the value vector $\boldsymbol{V}$. The attention relationship is established using $A(\boldsymbol{Q},\boldsymbol{K},\boldsymbol{V})=\text{softmax}(\boldsymbol{Q}\times \boldsymbol{K}^\top)\times \boldsymbol{V}$, where $\times$ represents matrix multiplication.
	To enable cross-frame attention, we process the feature via dual paths, as
	\begin{equation}
		\small
		\begin{aligned}
			&\boldsymbol{f}'_{d, t_1}=A(\boldsymbol{f}_{d, t_1},\boldsymbol{f}_{d, t_1},\boldsymbol{f}_{d, t_1}), \boldsymbol{f}''_{d, t_1}=A(\boldsymbol{f}_{d, t_1},\boldsymbol{f}_{d, t_2},\boldsymbol{f}_{d, t_2}),\\
			&\boldsymbol{f}'_{d, t_2}=A(\boldsymbol{f}_{d, t_2},\boldsymbol{f}_{d, t_2},\boldsymbol{f}_{d, t_2}), \boldsymbol{f}''_{d, t_2}=A(\boldsymbol{f}_{d, t_2},\boldsymbol{f}_{d, t_1},\boldsymbol{f}_{d, t_1}).
		\end{aligned}
	\end{equation}
	Moreover, a short-range fusion operation is set as the refinement operation to the propagation at both spatial and channel levels. This can be written as
	\begin{equation}
		\small
		\begin{aligned}
		\hat{\boldsymbol{f}}_{d, t_1},\hat{\boldsymbol{f}}_{d, t_2}=S(C(&(\boldsymbol{f}_{d, t_1}+\boldsymbol{f}'_{d, t_1}+\boldsymbol{f}''_{d, t_1})\oplus\\ &(\boldsymbol{f}_{d, t_2}+\boldsymbol{f}'_{d, t_2}+\boldsymbol{f}''_{d, t_2}))),
		\end{aligned}
	\end{equation}
	where $\oplus$ represents channel concatenation, $C$ refers to the convolution network, and $S$ signifies channel dimension splitting. The decomposition results for $\I_{d,t_1}$ and $\I_{d,t_2}$ are obtained by processing $\hat{\boldsymbol{f}}_{d, t_1}$ and $\hat{\boldsymbol{f}}_{d, t_2}$ through the decoder $M_D$. We have confirmed that our dual network with synchronous supervision for the outputs of $\I_{d,t_1}$ and $\I_{d,t_2}$ can yield superior decomposition and enhancement results compared to individual enhancement strategies.
	
 \minisection{The Role of the Cross-attention in CFIM}
	The cross-attention mechanism facilitates information propagation across different frames, aligning with our training procedure that requires spatial consistency for $\R$.  
	Moreover, during training, CFIM brings varying filtered features from randomly selected reference images ($\I_{d,t_2}$) into the backbone (i.e., the part without feature propagation), where the input is the image at the current time step $\I_{d,t_1}$. This setup enables CFIM to guide the backbone in learning how to adaptively leverage diverse knowledge from reference images, resulting in robust and effective feature spaces within the backbone, without the need for expensive temporal alignment.
	CFIM also leverages cross-frame information when it complements backbone.
	
	\subsection{Overall Objective to Compute Loss Function}
	During the training, we find it is better to adopt the dual training strategy, i.e., for each $\I_{d, t_1}$, we sample its neighboring reference $\I_{d, t_2}$ and constrain their outputs simultaneously.
	Thus, the loss function can be written as
	\begin{equation}
		\begin{aligned}
					\mathcal{O}_v=\mathbb{E}_{t_1, t_2} [ &\mathcal{O}_{rec, t_1} + \mathcal{O}_{rec, t_2} +\\ & \lambda_1 (\mathcal{O}(\L_{t_1})+\mathcal{O}(\L_{t_2}))+\lambda_2\mathcal{O}(\R_{t_1,t_2})],
		\end{aligned}
		\label{eq:video2}
	\end{equation}
	where $\lambda_1$ and $\lambda_2$ are the loss weights, and each loss term is defined in \Cref{eq:1}, \Cref{eq:2} and \Cref{eq:3}.

\section{Experiments}

\begin{table}[t]
	\centering
    \huge
	\resizebox{1.0\linewidth}{!}
    {
        \begin{tabular}{|l|cc|cc|cc|cc|}
            \hline
			& \multicolumn{2}{c|}{SDSD-indoor} &\multicolumn{2}{c|}{SDSD-outdoor}& \multicolumn{2}{c|}{SMID} &\multicolumn{2}{c|}{DID}  \\
            \hline
			Methods & PSNR & SSIM& PSNR & SSIM & PSNR & SSIM& PSNR & SSIM\\
			\hline \hline
			SNR    &27.30 &0.84 &23.23 & 0.82& 28.49 &0.81 &24.85 &0.90 \\
			SMG    &27.82 &0.86 & 25.17&0.81 &28.03 &0.78 &25.14 &0.85 \\
			PairLLE    &23.48 &0.71 &20.04 &0.65 &22.70 &0.63 &22.56 &0.82 \\
			RetinexFormer     & 26.56& 0.79 &22.80 &0.77 &29.15 &0.82 & 25.40&0.89 \\
			\hline
			MBLLEN & 22.17 & 0.66 &  21.41 & 0.63 & 22.67 & 0.68 & 24.22 & 0.86\\
			SMID & 24.84 & 0.72 & 23.30&0.67 & 24.78 &0.72 & 22.28&0.84\\
			SMOID & 24.63 & 0.70 & 22.25 & 0.68 & 23.64 & 0.71&22.13 &0.85\\
			SDSDNet& 26.81 & 0.75 & 23.08 & 0.71 & 26.03 &0.75 & 22.52 & 0.81\\
			DP3DF & 27.63 & 0.74 & 23.85 & 0.73 & 27.19 & 0.76 & 22.39 & 0.88\\
			StableLLVE & 25.32 & 0.70 & 22.47 & 0.65 & 24.37 & 0.69 & 21.64 & 0.80\\
			LLVE-SEG & 26.19 & 0.77 & 24.09 & 0.72 & 25.31 & 0.74 & 23.85 & 0.82\\
			LLVE-CFA & 24.28 &  0.81 & 22.64 & 0.76 & 24.72 & 0.70 & 22.53 & 0.87\\
			\hline
			BLLRVE&26.02 & 0.74&23.50&0.71&26.15&0.73	&23.25& 0.81\\
			VRGSS~&27.81 &0.76&24.28& 0.75&27.84&0.78&	24.51&0.87\\
			ZRLLE-PQP&25.53& 0.71&22.42&0.73&26.36&0.77	&22.84&0.80\\
			JDAE&26.19&0.74&23.37&0.76&26.45&0.76	&23.53&0.82\\
			\hline 
			Ours      & \textbf{28.93}&\textbf{0.88} &\textbf{26.32} & \textbf{0.82}& \textbf{29.60} &\textbf{0.82} & \textbf{30.10}&\textbf{0.93} \\
            \hline
	\end{tabular}}
    \caption{Quantitative comparison on SDSD, SMID, and DID datasets. Our method performs the best consistently.} 
	\label{comparison5}
\end{table}

\subsection{Datasets}
Our evaluation is conducted on four publicly available datasets, which encompass a wide range of real-world videos with diverse motion patterns and degradations, including SMID~\cite{chen2019seeing}, SDSD~\cite{wang2021sdsd}, DID~\cite{fu2023dancing}, and DAVIS~\cite{pont20172017}.

\subsection{Implementation Details}

\minisection{Experimental Details}
We conducted experiments on all datasets using the same network structure. 
$T$ is set as 5 in the experiment.
All modules were trained end-to-end, with the learning rate initialized at $4e^{-4}$ for all layers, adapted by a cosine learning scheduler. The batch size used was 4. Correspondences were computed using the SOTA method DKM~\cite{edstedt2023dkm}. 
We used the pre-trained weights of DKM during training because it is designed for general indoor and outdoor scenes. Its generalization ability is supported by its extensive training data and state-of-the-art training strategy, as demonstrated in the original DKM paper and subsequent studies~\cite{zhu2023pmatch,edstedt2024roma} (through evaluations on unseen scenes).

\subsection{Comparison and Baselines}
We conducted a comprehensive comparison with SOTA LLVE methods, including MBLLEN~\cite{lv2018mbllen}, SMID~\cite{chen2019seeing}, SMOID~\cite{jiang2019learning}, SDSDNet~\cite{wang2021sdsd}, DP3DF~\cite{xu2023deep}, StableLLVE~\cite{zhang2021learning}, LLVE-SEG~\cite{liu2023low},  LLVE-CFA~\cite{chhirolya2022low}, BLLRVE~\cite{zhang2024binarized}, VRGSS~\cite{li2023simple}, ZRLLE-PQP~\cite{wang2024zero}, and JDAE~\cite{shi2024zero}.
We also compared our method with SOTA LLIE methods for individual frames, including SNR~\cite{xu2022snr}, SMG~\cite{xu2023low}, PairLLE~\cite{fu2023learning}, RetinexFormer~\cite{cai2023retinexformer}. All methods were trained on each dataset using their respective released code and hyper-parameters (e.g., the training epoch that is set to guarantee convergence).
\textit{Note that all baselines are trained on our unified data split for a fair comparison. The splits slightly differ from those in baselines' original papers, so scores of baselines may vary from original papers}.

\begin{figure*}[t]
	\centering
	\begin{subfigure}[c]{0.16\textwidth}
		\centering
		\includegraphics[width=1.15in]{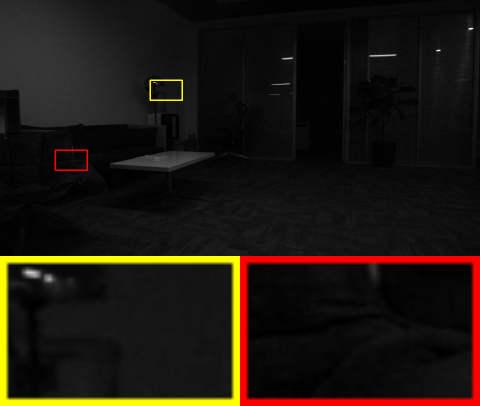}
	\end{subfigure}
	\begin{subfigure}[c]{0.16\textwidth}
		\centering
		\includegraphics[width=1.15in]{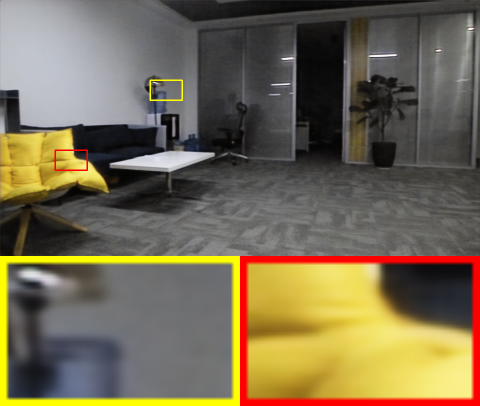}
	\end{subfigure}
	\begin{subfigure}[c]{0.16\textwidth}
		\centering
		\includegraphics[width=1.15in]{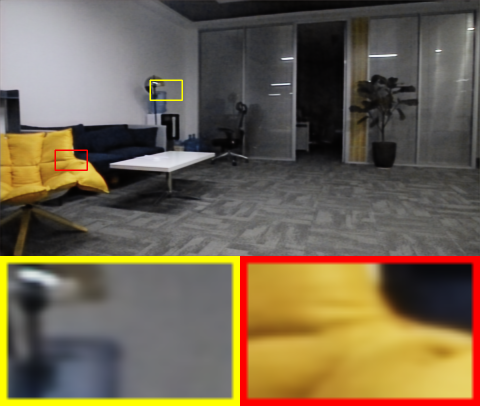}
	\end{subfigure}
	\begin{subfigure}[c]{0.16\textwidth}
		\centering
		\includegraphics[width=1.15in]{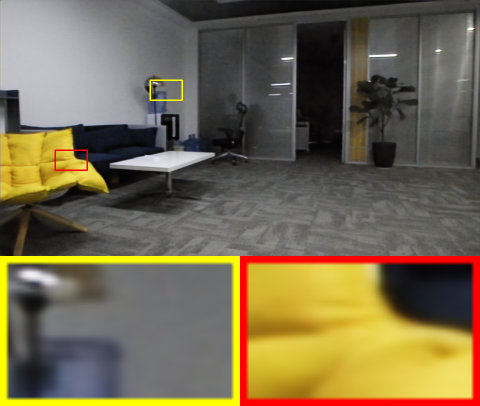}
	\end{subfigure}
	\begin{subfigure}[c]{0.16\textwidth}
		\centering
		\includegraphics[width=1.15in]{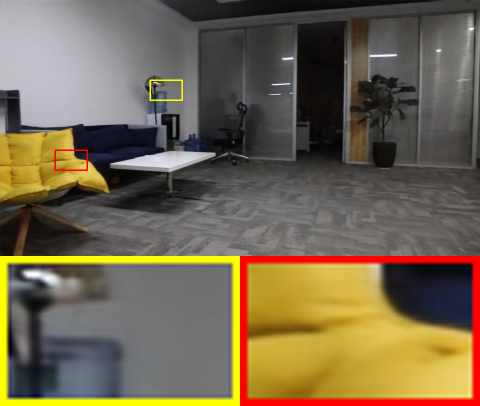}
	\end{subfigure} 
	\begin{subfigure}[c]{0.16\textwidth}
		\centering
		\includegraphics[width=1.15in]{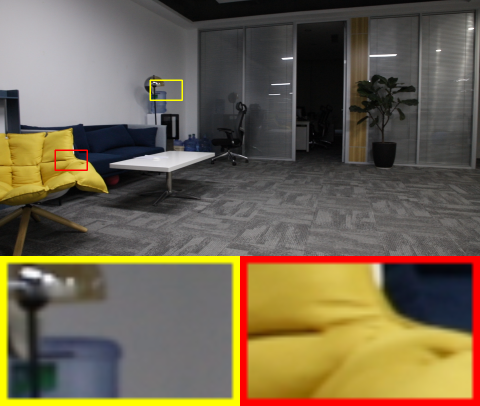}
	\end{subfigure} 
    \\
	\begin{subfigure}[c]{0.16\textwidth}
		\centering
		\includegraphics[width=1.15in]{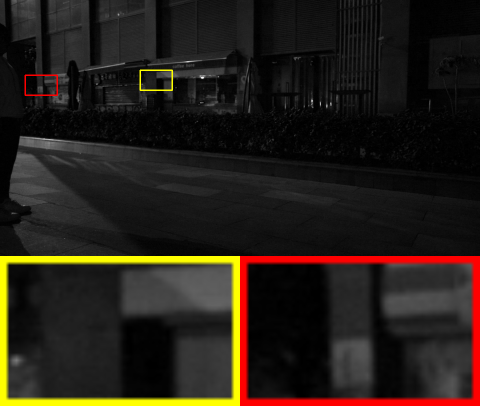}
	\end{subfigure}
	\begin{subfigure}[c]{0.16\textwidth}
		\centering
		\includegraphics[width=1.15in]{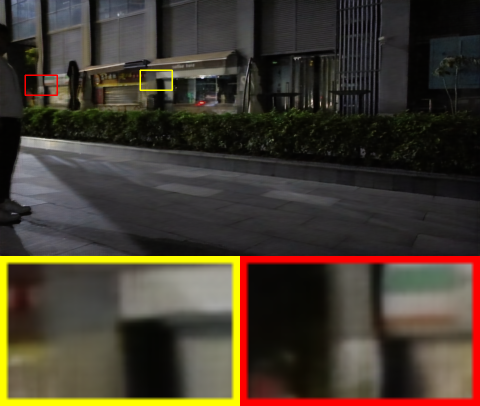}
	\end{subfigure}
	\begin{subfigure}[c]{0.16\textwidth}
		\centering
		\includegraphics[width=1.15in]{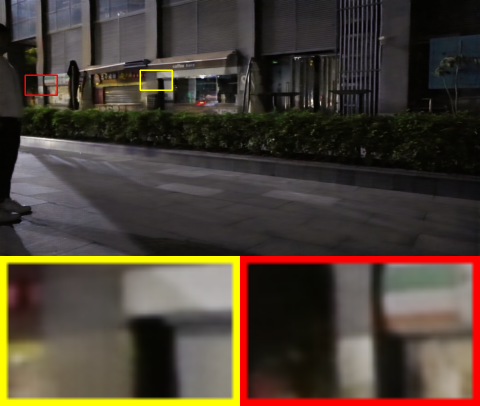}
	\end{subfigure}
	\begin{subfigure}[c]{0.16\textwidth}
		\centering
		\includegraphics[width=1.15in]{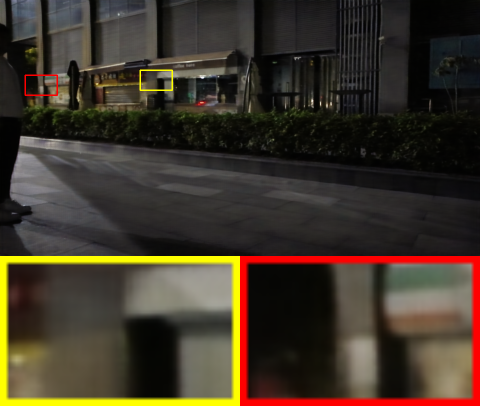}
	\end{subfigure}
	\begin{subfigure}[c]{0.16\textwidth}
		\centering
		\includegraphics[width=1.15in]{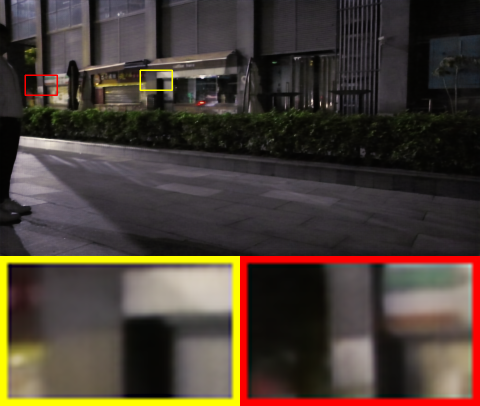}
	\end{subfigure} 
	\begin{subfigure}[c]{0.16\textwidth}
		\centering
		\includegraphics[width=1.15in]{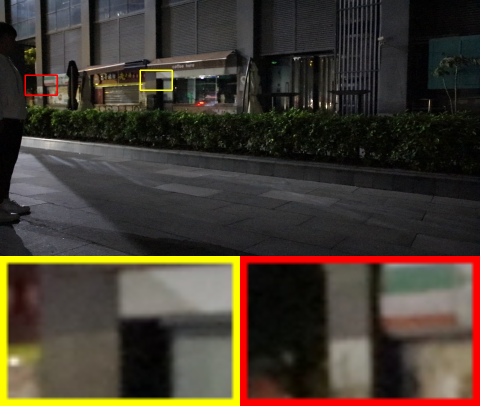}
	\end{subfigure} 
    \\
	
	\begin{subfigure}[c]{0.16\textwidth}
		\centering
		\includegraphics[width=1.15in]{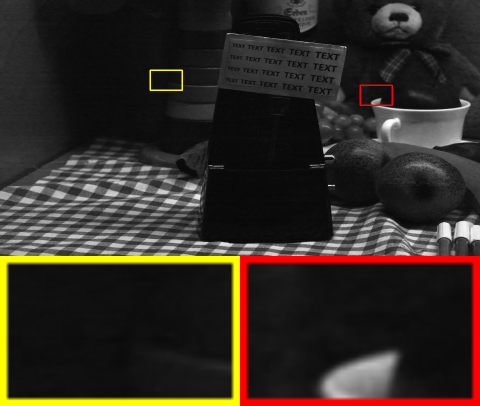}
		\caption*{Input}
	\end{subfigure}
	\begin{subfigure}[c]{0.16\textwidth}
		\centering
		\includegraphics[width=1.15in]{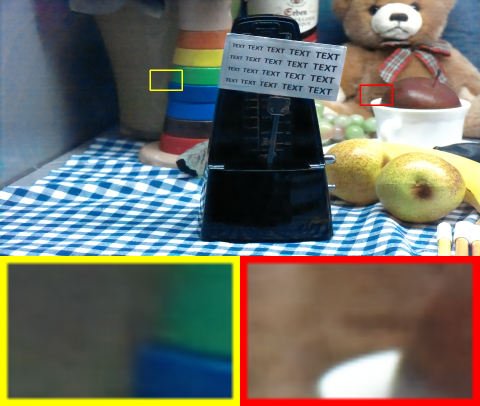}
		\caption*{RetinexFormer}
	\end{subfigure}
	\begin{subfigure}[c]{0.16\textwidth}
		\centering
		\includegraphics[width=1.15in]{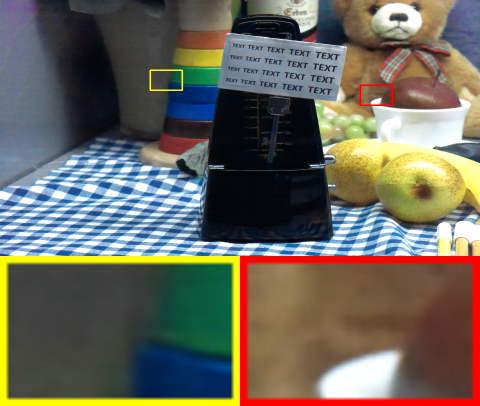}
		\caption*{SDSDNet}
	\end{subfigure}
	\begin{subfigure}[c]{0.16\textwidth}
		\centering
		\includegraphics[width=1.15in]{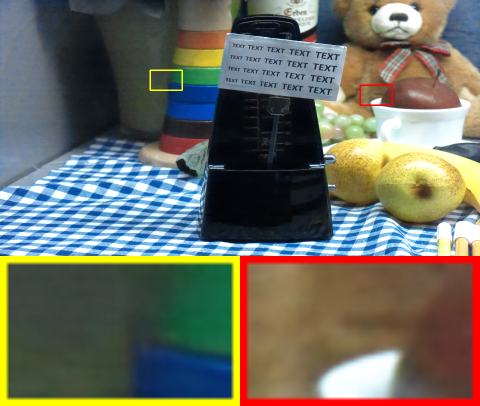}
		\caption*{DP3DF}
	\end{subfigure}
	\begin{subfigure}[c]{0.16\textwidth}
		\centering
		\includegraphics[width=1.15in]{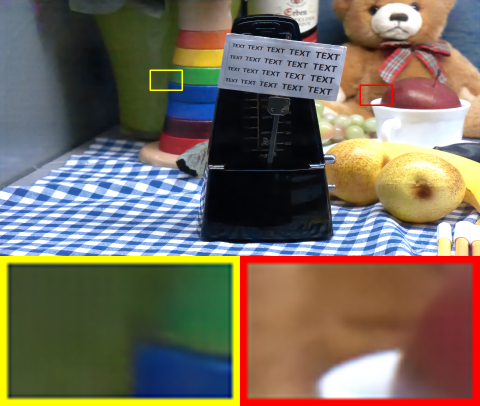}
		\caption*{Ours}
	\end{subfigure} 
	\begin{subfigure}[c]{0.16\textwidth}
		\centering
		\includegraphics[width=1.15in]{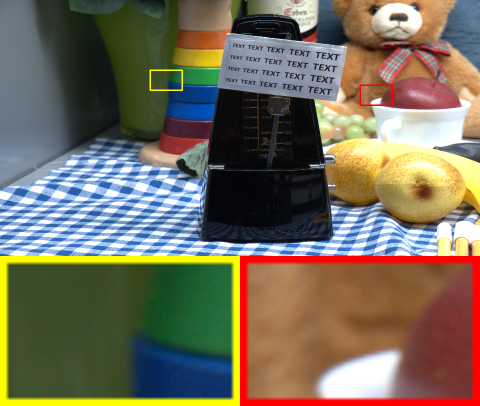}
		\caption*{GT}
	\end{subfigure}
    \\
	\caption{Visual comparisons on SDSD-indoor (top two rows), SDSD-outdoor (middle two rows), and SMID (bottom two rows). The results of our proposed framework, i.e., ``Ours'', demonstrate better visual perception with clearer visibility and more enhanced details.}
	\label{fig:cmp1}
\end{figure*}

\begin{figure*}[t]
	\centering
	\begin{subfigure}[c]{0.16\textwidth}
		\centering
		\includegraphics[width=1.15in]{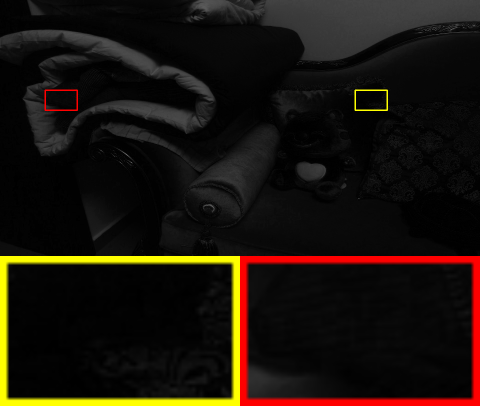}
	\end{subfigure}
	\begin{subfigure}[c]{0.16\textwidth}
		\centering
		\includegraphics[width=1.15in]{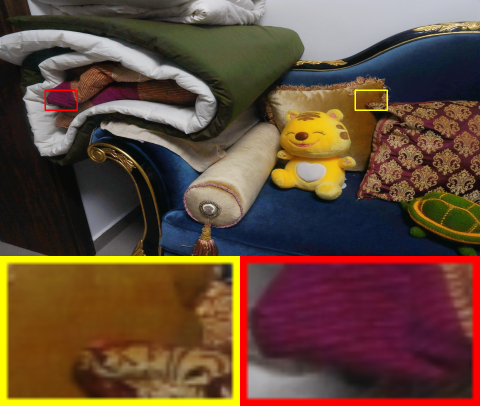}
	\end{subfigure}
	\begin{subfigure}[c]{0.16\textwidth}
		\centering
		\includegraphics[width=1.15in]{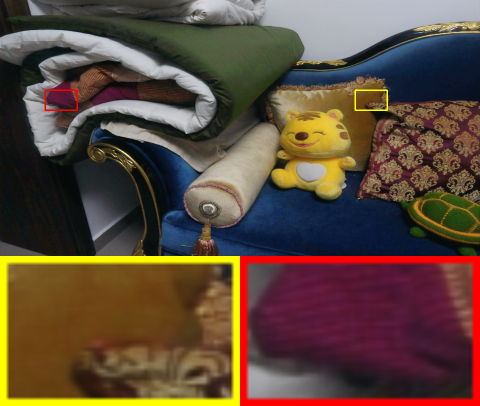}
	\end{subfigure}
	\begin{subfigure}[c]{0.16\textwidth}
		\centering
		\includegraphics[width=1.15in]{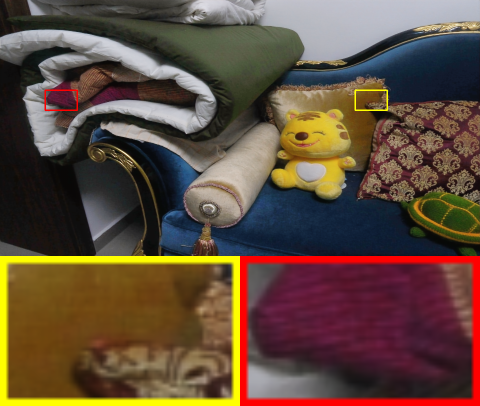}
	\end{subfigure}
	\begin{subfigure}[c]{0.16\textwidth}
		\centering
		\includegraphics[width=1.15in]{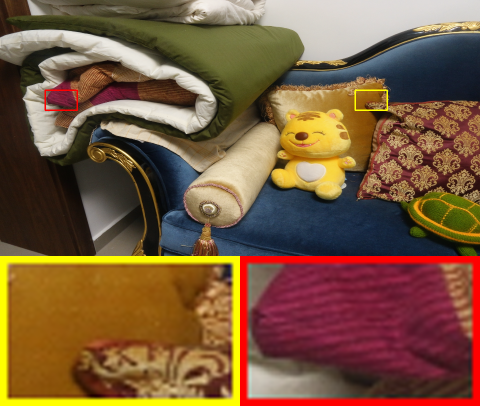}
	\end{subfigure} 
	\begin{subfigure}[c]{0.16\textwidth}
		\centering
		\includegraphics[width=1.15in]{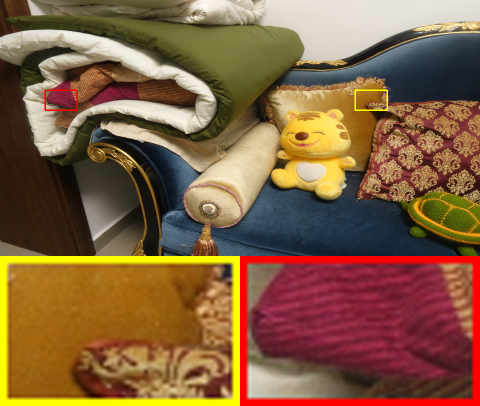}
	\end{subfigure} 
    \\

	\begin{subfigure}[c]{0.16\textwidth}
		\centering
		\includegraphics[width=1.15in]{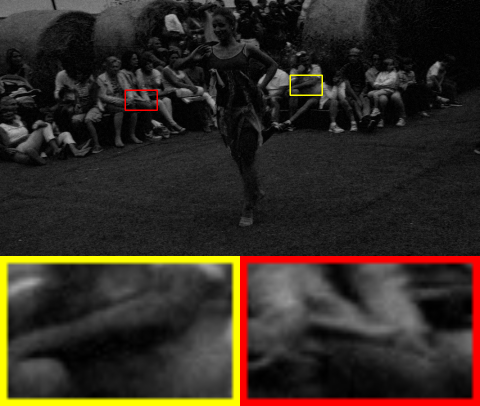}
		\caption*{Input}
	\end{subfigure}
	\begin{subfigure}[c]{0.16\textwidth}
		\centering
		\includegraphics[width=1.15in]{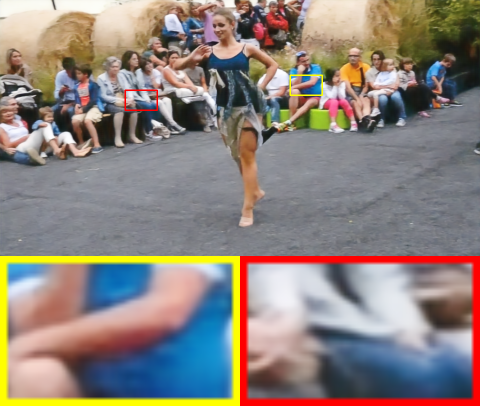}
		\caption*{RetinexFormer}
	\end{subfigure}
	\begin{subfigure}[c]{0.16\textwidth}
		\centering
		\includegraphics[width=1.15in]{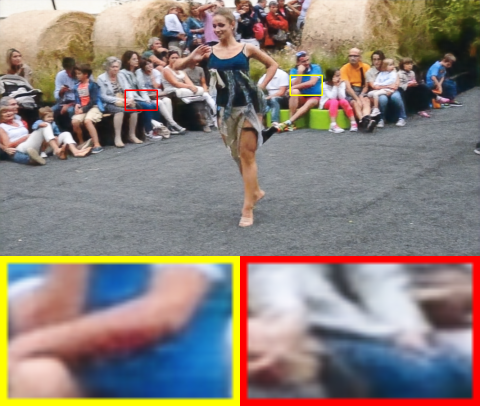}
		\caption*{SDSDNet}
	\end{subfigure}
	\begin{subfigure}[c]{0.16\textwidth}
		\centering
		\includegraphics[width=1.15in]{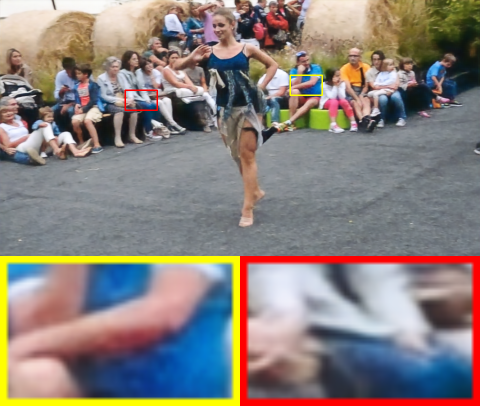}
		\caption*{DP3DF}
	\end{subfigure}
	\begin{subfigure}[c]{0.16\textwidth}
		\centering
		\includegraphics[width=1.15in]{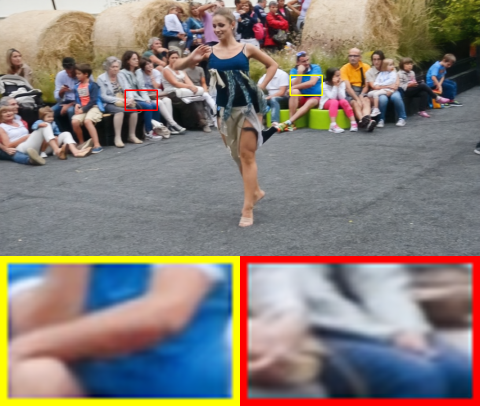}
		\caption*{Ours}
	\end{subfigure} 
	\begin{subfigure}[c]{0.16\textwidth}
		\centering
		\includegraphics[width=1.15in]{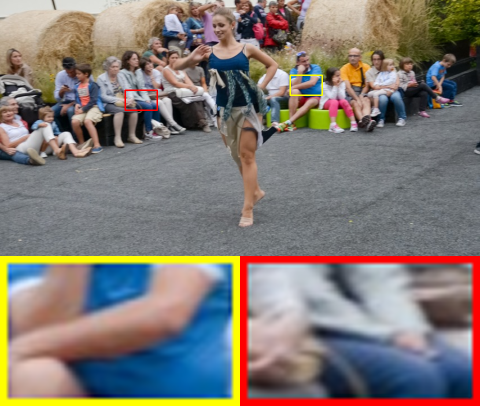}
		\caption*{GT}
	\end{subfigure} 
    \\
	\caption{Comparisons on DID (top two rows) and DAVIS (bottom two rows). ``Ours'' has better visibility and details.}
	\label{fig:cmp2}
\end{figure*}

\begin{table}[tb!]
	\centering
	\huge
	\resizebox{1.02\linewidth}{!}
    {
		\begin{tabular}{|l|cccccc|}
            \hline
			&SNR&SMG&PairLLE&RetinexFormer&MBLLEN&SMID\\
			\hline \hline
			PSNR &20.69 &20.18 & 19.57&21.79 & 18.63&20.51\\
			SSIM&0.710 & 0.672&0.667 &0.723 &0.619 & 0.686\\
			\hline \hline
			&SMOID&SDSDNet&DP3DF&StableLLVE&LLVE-SEG&Ours\\
			\hline
			PSNR &20.66 & 21.22&22.04 &21.48 &21.45 &\textbf{23.38}\\
			SSIM&0.697 &0.716 & 0.749&0.732 &0.715 &\textbf{0.782} \\
            \hline
	\end{tabular}}
    \caption{Quantitative comparison on the DAVIS dataset.}
	\label{comparison2}
\end{table}

\minisection{Quantitative Result}
In \Cref{comparison5}, we present the comparative results with the selected baseline methods across the SDSD, SMID, and DID datasets.
The table reveals that our approach {\em consistently\/} outperforms all other methods, as indicated by the highest PSNR and SSIM scores on all datasets. Notably, our scores exhibit a substantial lead over all others, particularly on DID, which is a large-scale dynamic video dataset. This superiority underscores the robust capability of our method in enhancing real-world videos.

Furthermore, \Cref{comparison2} provides a summary of the comparative results on DAVIS. In comparison to the degradation synthesis strategy as presented in \cite{zhang2021learning}, we have extended our approach to include the degradation of a low-light noise term. As a result, our synthesized data encompasses dynamic scenes with pronounced invisibility and perturbations, posing a considerable challenge. As illustrated in \Cref{comparison2}, our method consistently yields the highest PSNR and SSIM values, reaffirming the effectiveness of our approach.

\begin{table}[tb!]
	\centering
	\huge
    \renewcommand{\arraystretch}{1.06}
	\resizebox{1.0\linewidth}{!}
    {
		\begin{tabular}{|l|cc|cc|cc|cc|}
            \hline
			& \multicolumn{2}{c|}{DAVIS} &\multicolumn{2}{c|}{SDSD-Indoor}&\multicolumn{2}{c|}{SDSD-Outdoor}& \multicolumn{2}{c|}{DID}\\
			\hline 
			Methods & Short & Long& Short & Long& Short & Long& Short & Long \\
			\hline \hline
			SNR &0.027 & 0.070& 0.017&0.060 &0.022 &0.046 & 0.025&0.068 \\
			RetinexFormer &0.029 &0.072 &0.018 &0.061 &0.024 &0.043 & 0.024&0.065 \\
			SDSDNet &0.024 &0.068 & 0.012&0.053 &0.011 &0.038 &0.017 &0.059 \\
			DP3DF & 0.021 &0.065 &\textbf{0.011} &\textbf{0.050} &0.013 & 0.036&0.019 &0.062 \\
			StableLLVE &0.023 &0.067 &0.016 &0.055 &0.019 &0.042 &0.023 &0.066 \\
			LLVE-SEG &0.025 &0.071 &0.014 & 0.057&0.016 &0.040 & 0.021&0.064 \\
			\hline \hline
			Ours-R &\textbf{0.017} &\textbf{0.058} &\textbf{0.010}&\textbf{0.048}&\textbf{0.007}&\textbf{0.030} &\textbf{0.011} &\textbf{0.049}  \\
			
			Ours &\textbf{0.020} &\textbf{0.063} &\textbf{0.012}&\textbf{0.051}&\textbf{0.010}&\textbf{0.034} &\textbf{0.015} &\textbf{0.054}  \\
            \hline
	\end{tabular}}
    	\caption{The quantitative comparison in terms of short-term (``Short'') and long-term (``Long'') temporal loss. ``Ours-R'' means the view-independent term produced by ours.}
	\label{comparison5-temporal}
\end{table}

\minisection{Evaluation for Temporal Consistency}
For video processing, the performance of temporal consistency and stability should be evaluated. 
Thus, we employ the short-term and long-term temporal loss proposed in \cite{lai2018learning} for such temporal evaluation on different datasets.
The wrapping operations among frames are computed on the frames with the normal light.
Moreover, the long-term loss is computed every 10 frames.
The results are shown in \Cref{comparison5-temporal} (we normalize the frame values into [0, 1]). 
Obviously, our results have lower temporal loss than the baselines, e.g., higher temporal consistency and stability.
Specifically, we assess the temporal consistency of the view-independent term. The results presented in \Cref{comparison5-temporal} further validate the constancy.

\minisection{Qualitative Result}
Besides the quantitative comparison, we present visual comparisons with the selected baselines.
\Cref{fig:cmp1} showcases the visual comparisons of SDSD and SMID, while \Cref{fig:cmp2} displays visual cases from DID and DAVIS. In general, the results enhanced by our approach exhibit a more natural appearance, including accurate color, well-balanced brightness, enhanced contrast, and precise details.
Our results show fewer artifacts in regions with complex textures, and they appear cleaner and sharper compared to results produced by other methods. This distinction is particularly notable, as most baselines perform well in simpler areas.

\begin{table}[tb!]
	\centering
    \huge
    \renewcommand{\arraystretch}{1.06}
	\resizebox{1.0\linewidth}{!}
    {
		\begin{tabular}{|l|cc|cc|cc|cc|}
            \hline
			& \multicolumn{2}{c|}{SDSD-indoor} &\multicolumn{2}{c|}{SDSD-outdoor}& \multicolumn{2}{c|}{SMID} &\multicolumn{2}{c|}{DID}  \\ \hline
			Methods & PSNR & SSIM& PSNR & SSIM & PSNR & SSIM& PSNR & SSIM\\
			\hline \hline
			w/o LR & 25.78 &0.77  &23.49  & 0.75 &26.37 &0.76  & 27.54 &0.86 \\
			w/o LL & 26.89 & 0.80 & 24.19 &0.78  &27.74 & 0.79 & 27.72 & 0.89\\
			w/o C.F. & 25.18 & 0.82 & 24.22 & 0.77 &26.75 &0.77  &25.60  &0.85 \\
			with M.I. & 26.21 & 0.84 & 25.36 & 0.80 &28.42 &0.79  &28.33  &0.88 \\
			w/o Dual & 27.53 & 0.84 &24.83  &0.79  &27.43 & 0.78 & 26.91 &0.86 \\
			\hline 
			Full      & \textbf{28.93}&\textbf{0.88} &\textbf{26.32} & \textbf{0.82}& \textbf{29.60} &\textbf{0.82} & \textbf{30.10}&\textbf{0.93} \\
            \hline
	\end{tabular}}
    \caption{Ablation study on SDSD, SMID, and DID.} 
    \label{comparison-abla}
\end{table}

\begin{figure*}[!t]
	\begin{center} 
		\includegraphics[width=0.9\linewidth]{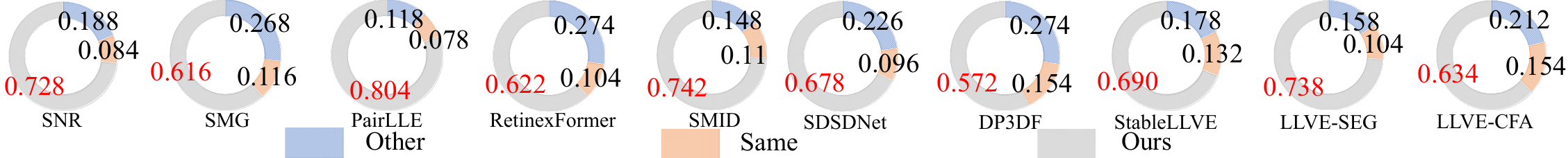}
	\end{center}
	\caption{
		The above pie charts summarize the results of our user study, and ours is preferred by participants.
	}
	\label{us_tbl}
\end{figure*}

\subsection{Ablation Study}
\label{sec:ablation-study}

We assess the critical components of our framework through five ablation cases. (1) ``w/o LR'' that removes the constraint on learning the view-independent part. (2) ``w/o LL'' where the constraint on learning the view-dependent part is removed. (3) ``w/o C.F.'': we eliminate the cross-frame attention and fusion in the network, resulting in each frame being enhanced individually. (4) ``with M.I.'': This refers to using multiple neighboring frames as input and employing the temporal alignment of deformable convolution. (5) ``w/o Dual'': the dual learning strategy is removed, meaning that the loss is applied to only one frame in each iteration.

The results are summarized in \Cref{comparison-abla}. By comparing ``w/o LR'' with the full setting (``Full''), we can clearly demonstrate the effectiveness of our proposed constraint for the view-independent part. Similarly, the significant advantage of ``Full'' over ``w/o LL'' highlights the significance of the proposed continuity constraint. 
Furthermore, we can validate the importance of mutually propagating features among different frames to achieve spatial-temporal consistent decomposition by comparing ``w/o C.F.'' with ``Full''. Our proposed propagation strategy via the dual network structure proves to be more effective than the common method of using multi-frame inputs for propagation in videos, as seen in the comparison between ``w/o M.I.'' and ``Full''. 
The dual learning setting, where we simultaneously supervise the two outputs of the dual network, also proves to be valuable, as evidenced by the comparison between ``w/o Dual'' and ``Full''.

\subsection{User Study}
To prove the effectiveness of our proposed framework in terms of human subjective evaluation, we conduct a large-scale user study with 100 participants with varying ages and education backgrounds, and balanced sex distribution. 
Following most of the existing low-light enhancement works~\cite{xu2023low,wang2023lighting}, we employ the AB-test for the user study. 
Participants should indicate their preference or select the same option.
They should make the decision according to the natural brightness, contrast, and color of each frame; the rich details, fewer artifacts, and the temporal consistency in the video. 

\Cref{us_tbl} summarizes the user study's results, and we can see that ours gets more selections from participants over all the baselines. This demonstrates that our method's results are more preferred by the human subjective perception.

\subsection{The Effects of CFIM Structure}
As previously discussed, the use of CFIM yields two key benefits: it enhances the consistency of the image enhancement process and strengthens the robustness of the backbone by providing the cross-attention mechanism with information from random neighboring time steps.
In this section, we conduct experiments to show the effects of CFIM.
The baseline consists of a network that processes a single frame, while the comparative setup integrates CFIM, where cross-attention is applied between randomly selected pairs of frames during training. The results of this comparison are presented in Table~\ref{comparison-abla-supp-cfim}, which show that models incorporating the CFIM structure significantly outperform those without it, thereby demonstrating the effectiveness of the CFIM approach.
Besides, the ablation results of the mentioned ``w/o C.F.'' (in Sec.~\ref{sec:ablation-study}) also support the effects of CFIM in the decomposition.

\begin{table}[tb!]
	\centering
	\huge
    \renewcommand{\arraystretch}{1.06}
    \resizebox{1.0\linewidth}{!}
    {
        \begin{tabular}{|l|cc|cc|cc|cc|}
            \hline
            & \multicolumn{2}{c|}{SDSD-indoor} &\multicolumn{2}{c|}{SDSD-outdoor}& \multicolumn{2}{c|}{SMID} &\multicolumn{2}{c|}{DID}  \\
            \hline
            Methods & PSNR & SSIM& PSNR & SSIM & PSNR & SSIM& PSNR & SSIM\\
            \hline \hline
            SNR &27.30 &0.84 &23.23 & 0.82& 28.49 &0.81 &24.85 &0.90\\
            SNR+CFIM &\textbf{28.15} &\textbf{0.86} &\textbf{24.57} &\textbf{0.83} &\textbf{29.24} &\textbf{0.82} &\textbf{26.03} &\textbf{0.91}\\ \hline
            R.F.& 26.56& 0.79 &22.80 &0.77 &29.15 &0.82 & 25.40&0.89\\
            R.F.+CFIM &\textbf{27.42} &\textbf{0.80} &\textbf{23.75} &\textbf{0.79} &\textbf{29.87} &\textbf{0.83} &\textbf{26.38} &\textbf{0.90}\\ \hline
        \end{tabular}
        }
        \caption{The effects of CFIM. ``R.F." denotes RetinexFormer.} 
    \label{comparison-abla-supp-cfim}
\end{table}

\subsection{Robustness Towards Correspondences}
\label{sec:robust}
In this section, we explore the robustness of our framework with respect to incorrect and insufficient correspondences.
In the first experiment, we introduce random noise perturbations to the computed correspondences, modifying the location values of the correspondences (the perturbation range is sampled from $-20 \sim 20$ pixels, which falls within the normal error range of current correspondence estimation methods~\cite{edstedt2023dkm}). As shown in Table~\ref{comparison-abla-supp-robust}, the results with perturbed correspondences (denoted as ``Ours with per.") show a performance drop compared to the unperturbed case. However, the performance remains superior to that of most SOTA baselines, indicating that our framework is robust to a certain of erroneous correspondences, which are likely to occur in areas with complex textures or other challenging scenarios.
In the second experiment, we address the issue of insufficient correspondences, which can arise, for instance, in regions with occlusions. To simulate this, we randomly reduce the number of correspondences to 10\% of the originally detected ones. The results (``Ours with red.") are presented in Table~\ref{comparison-abla-supp-robust}, where we observe that our method continues to outperform almost all baselines in Table~\ref{comparison5}, albeit with a smaller margin than the original setting. This suggests that even with fewer correspondences, the remaining ones still provide valuable information and they are in accord with the other constraints. This further demonstrates the robustness of our method.

\begin{table}[tb!]
	\centering
	\huge
    \renewcommand{\arraystretch}{1.06}
    \resizebox{1.0\linewidth}{!}
    {
        \begin{tabular}{|l|cc|cc|cc|cc|}
            \hline
            & \multicolumn{2}{c|}{SDSD-indoor} &\multicolumn{2}{c|}{SDSD-outdoor}& \multicolumn{2}{c|}{SMID} &\multicolumn{2}{c|}{DID}  \\
            \hline
            Methods & PSNR & SSIM& PSNR & SSIM & PSNR & SSIM& PSNR & SSIM\\
            \hline
            Ours with red. &27.97 &0.84 &25.73 & 0.80&29.12 &0.79 & 28.86&0.89\\
            Ours with per. &28.10 &0.86 &26.04 &0.79 &29.37 &0.81 &29.08 &0.91\\
            Ours      & 28.93&0.88 &26.32 & 0.82& 29.60 &0.82 & 30.10&0.93 \\
            \hline
        \end{tabular}
        }
        \caption{The robustness of our framework towards correspondences.} 
    \label{comparison-abla-supp-robust}
\end{table}

\section{Conclusion}
We introduce a novel LLVE framework that utilizes spatial-temporal consistent decomposition and dual networks for mutual feature propagation. It predicts the decomposition of normal-light outputs without external priors, thanks to cross-frame correspondences and practical constraint of the continuity. 
Extensive experiments on various datasets showcase the framework's superiority over SOTA approaches and the impact of our designed components.

\section*{Ethical Statement}

There are no ethical issues.

\section*{Acknowledgments}

This work is supported by the Natural Science Foundation of Zhejiang Province, China, under No. LD24F020002.

\bibliographystyle{named}
\bibliography{ijcai25}

\end{document}